# Development of a Handheld Actuation Mechanism for a Tendon-driven Robotically Steered Guidewire

Saima Chavenet, Timothy A. Brumfiel, Revanth Konda, Jaydev P. Desai

*Medical Robotics and Automation (Robomed) Laboratory*

*Wallace H. Coulter Department of Biomedical Engineering, Georgia Institute of Technology*

*schavenet3@gatech.edu, tbrumfiel4@gatech.edu, rkonda30@gatech.edu, jaydev@gatech.edu*

## INTRODUCTION

Cardiovascular diseases (CVDs), which are conditions related to the blood vessels and the heart, are the leading cause of death worldwide, with estimates of about 19.41 million deaths attributed to CVDs in 2021 [1]. Endovascular interventions are minimally invasive procedures used to treat CVDs. An endovascular intervention begins with a skilled clinician manually navigating a long, slender wire, called a guidewire, to the target location within the vasculature [2]. Afterwards, the guidewire is used to place other medical devices, such as catheters and stents, to complete a given procedure. Due to factors, such as vessel tortuosity and lack of steerability at the guidewire tip, manual navigation of a guidewire could be challenging, potentially resulting in vessel damage, perforation, dissection, and occlusion, as well as post-surgical complications, such as thrombosis [3]. Robotically steerable guidewires, which are slender continuum robots with the ability to exhibit controlled bending motion at the tip [2], have been developed to alleviate the aforementioned challenges and improve navigation efficiency within the vasculature. This work focuses on tendon-driven robotically steerable guidewires, which exhibit advantageous properties such as the ability to generate sufficient actuation ranges and forces while being relatively compact [4], compared to magnetic actuation strategies [5].

Our previous work presented the design of a compact spooling mechanism capable of storing and dispensing clinically relevant lengths of tendon-driven robotically steerable guidewires [6]. However, the aforementioned mechanism relied on an external motorized system for roll motion and had a non-ergonomic form factor. This paper details the development of a handheld actuation mechanism for tendon-driven robotic guidewires by utilizing the aforementioned spooling mechanism. Incorporating the compact spooling mechanism into a handheld device enables clinicians to leverage both motorized guidewire steering and manual rotation and translation, if desired. The proposed device is designed such that, while holding the device, the clinician can execute feeding, rotation, and bending motions for up to 1.5 m of the robotically steerable guidewire using a joystick and momentary switch on the handle. Furthermore, the proposed device is demonstrated through navigation in an anatomically accurate phantom aorta model.

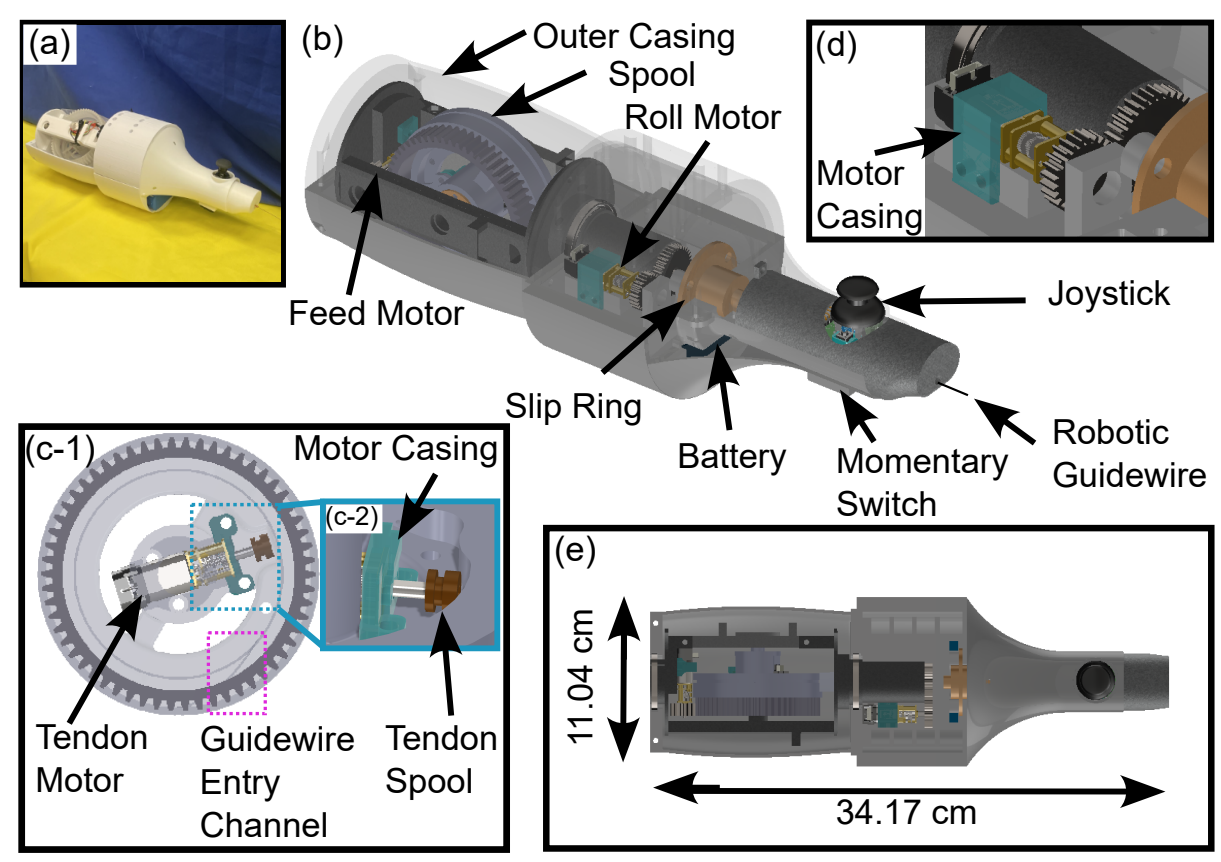

**Fig. 1** (a) An image of the handheld actuation mechanism developed in this work. A schematic of (b) the handheld actuation mechanism, (c-1) spooling mechanism, (c-2) tendon actuation motor, (d) roll assembly, and (e) a top view of the mechanism.

## MATERIALS AND METHODS

An image and schematic of the proposed device are shown in Fig. 1(a) and Fig. 1(b), respectively. The handheld actuation mechanism developed in this work is designed to accommodate a robotic guidewire with a length of up to 1.5 m and has a total weight of 0.836 kg. The bending segment of the robotic guidewire is fabricated by machining a 20 mm long unidirectional asymmetric notch (UAN) pattern into a 0.48 mm OD nitinol tube using a femtosecond laser (WS-Flex Ultra-Short Pulse Laser Workstation, Optec, Frameries, Belgium). The length of the bending segment was 20 mm. A nitinol tendon (76 $\mu$m) is attached to the distal tip of the guidewire using epoxy (J-B Weld, Sulphur Springs, TX, USA). Applying tension to the tendon facilitates actuation of the bending joint. The robotic guidewire is secured within the spool by a single, accessible screw. The guidewire is routed within a groove in the spool (Fig 1(c-1)) and is continuously wrapped on top of itself when retracted. Once the desired location is reached, the screw can be loosened and the guidewire can be removed from the mechanism, allowing a catheter to be placed over the guidewire. Within the spool, the tendon is attached to a smaller spool, as shown in Fig. 1(c-2), which can rotate to control the bending of the guidewire tip. The tendon spool has an OD of 7.4 mm and a groove with an OD of 5 mm. The spool features an embedded gear design, as shown in Fig. 1(b), which is rotated using a motor equipped with a driving gear, to control the feeding and retraction of the guidewire. The gear ratio between the spool and its driver gear is 29:6. The spool is attached in the center of a supporting brace, which can rotate. The mechanism to enable rotation of the brace is present at the front end of the device as shown in Fig. 1(b). A polypropylene strip was placed above the spool to ensure the guidewire remains within the groove during dispensing and retraction. The neck of the brace has an embedded gear at the distal end to control of the rotation of the guidewire, as shown in Fig. 1(d). The gear ratio between the brace gear and its driver gear is 34:25. All motors (Pololu Robotics and Electronics, NV, United States) are powered by a 12 V Li-ion battery, which is stored below the rotation mechanism at the distal end of the device. Slip rings are placed at several locations throughout the device to enable continuous rotational motion of each motor. The motors within the device are controlled by a Teensy 4.0, which, along with a PCB equipped with motor drivers, is placed above the rotation mechanism.

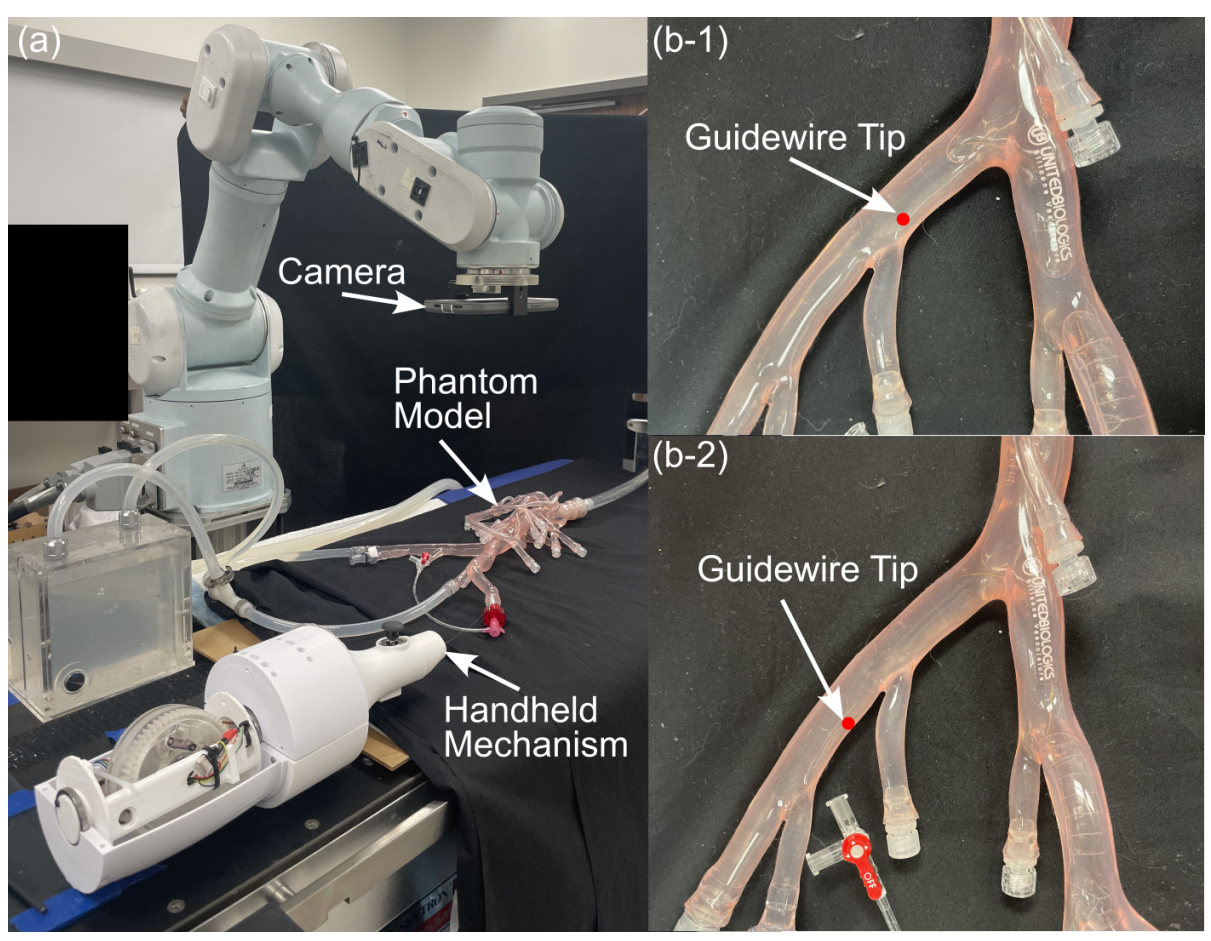

**Fig. 2** An image of (a) the setup for the navigation experiment in a phantom aorta model, (b-1) the guidewire crossing the aortic bifurcation, and (b-2) the guidewire moving down the right common iliac artery.

The wiring of the device extends into the handle portion of the design, which comprises an ergonomic design for clinicians to potentially use the device with comfort. The handle has an OD of 40 mm and houses the joystick and button controls for generating the different motions of the guidewire. Moving the joystick towards the distal or proximal end of the device enables the dispensing and retraction of the guidewire, respectively. Lateral motion of the joystick controls the rotation of the guidewire. The switch used to control the bending joint is a two-state momentary rocker switch. Pressing backward bends the guidewire through rotation of the tendon spool within the main spool, and pressing forward unbends the guidewire. The guidewire is passed from the spool through the center of the device through a polyimide tube (Zeus Company LLC, SC, USA). A top view of the device with relevant dimensions is presented in Fig. 1(e).

To demonstrate the use of the handheld actuation mechanism developed in this work, the robotic guidewire was navigated through an abdominal aorta model (United Biologics Inc., CA, USA). The experimental setup for this demonstration is shown in Fig. 2(a), where a camera was mounted above the phantom to image the guidewire during the navigation. The guidewire was introduced through the left femoral artery and navigated toward the femoral bifurcation. The guidewire was then bent across the femoral bifurcation, as shown in Fig. 2(b-1), and advanced further down the right iliac artery, as shown in Fig. 2(b-2). The experiment was qualitatively assessed. Similar to previously developed mechanisms [7], the guidewire was successfully navigated throughout the phantom. However, the proposed mechanism was capable of dispensing the guidewire without the tube buckling between the phantom insertion point and the mechanism exit, unlike our previous work [7], where required manual intervention was a critical shortcoming.

## DISCUSSION

The development of a handheld actuation mechanism for tendon-driven robotic guidewires was presented in this work. The device houses a spool and stores similar amounts of guidewire as previous compact spool designs [6]. The development of the handheld device enables efficient guidewire navigation within phantom models without requiring manual intervention. In our future work, we will modify the design to increase the clinical viability of this mechanism with additional considerations, such as guidewire replacement, sterilization, and user comfort.

## ACKNOWLEDGEMENT

Research reported in this publication was supported in part by the National Heart, Lung, And Blood Institute of the National Institutes of Health under Award Number R01HL144714. The content is solely the responsibility of the authors and does not necessarily represent the official views of the National Institutes of Health.